\pdfoutput=1
\documentclass[11pt]{article}
\usepackage[final]{acl}
\usepackage{times}
\usepackage{latexsym}
\usepackage[T1]{fontenc}
\usepackage[utf8]{inputenc}
\usepackage{microtype}
\usepackage{inconsolata}
\usepackage{bm}
\usepackage{makecell}
\usepackage{graphicx}
\usepackage{enumitem}
\usepackage{amsmath}
\usepackage{booktabs}
\usepackage{array}
\usepackage{graphicx}
\usepackage{multirow}
\usepackage{bbding}
\usepackage{colortbl}
\definecolor{mygray}{gray}{.92}
\usepackage[ruled,vlined]{algorithm2e}
 
\usepackage{amsmath,amsthm,amssymb}

\usepackage{color}
\title{Modality-Aware Integration  with Large Language Models\\for Knowledge-based Visual Question Answering}
\author{
    Junnan Dong\textsuperscript{1}, Qinggang Zhang\textsuperscript{1}, Huachi Zhou\textsuperscript{1}\\ \textbf{Daochen Zha\textsuperscript{2}, Pai Zheng\textsuperscript{1} Xiao Huang\textsuperscript{1}\thanks{~~Corresponding author}} \\
    \textsuperscript{1}The Hong Kong Polytechnic University, \textsuperscript{2} Rice University\\
    \{hanson.dong, qinggangg.zhang, huachi.zhou\}@connect.polyu.hk\\
    daochen.zha@rice.edu; \{pai.zheng, xiaohuang\}@polyu.edu.hk
}

\begin{document}
\maketitle
\begin{abstract}
Knowledge-based visual question answering (KVQA) has been extensively studied to answer visual questions with external knowledge, e.g., knowledge graphs (KGs). While several attempts have been proposed to leverage large language models (LLMs) as an implicit knowledge source, it remains challenging since LLMs may generate hallucinations. Moreover, multiple knowledge sources, e.g., images, KGs and LLMs, cannot be readily aligned for complex scenarios. To tackle these, we present a novel \underline{m}odality-\underline{a}ware \underline{i}ntegration with \underline{L}LMs for KVQA (\texttt{MAIL}). It carefully leverages multimodal knowledge for both image understanding and knowledge reasoning. Specifically, \((i)\) we propose a two-stage prompting strategy with LLMs to densely embody the image into a \textit{scene graph} with detailed visual features; \((ii)\) We construct a coupled \textit{concept graph} by linking the mentioned entities with external facts. \((iii)\) A tailored pseudo-siamese graph medium fusion is designed for sufficient multimodal fusion. 
We utilize the shared mentioned entities in two graphs as mediums to bridge a tight inter-modal exchange, while maximally preserving insightful intra-modal learning by constraining the fusion within mediums. Extensive experiments on two benchmark datasets show the superiority of \texttt{MAIL} with 24$\times$ less resources.

\end{abstract}

\section{Introduction}
Knowledge-based visual question answering (KVQA) aims to provide appropriate answers for questions about images based on external knowledge~\cite{wang2017fvqa,hong2024knowledgetosql}, such as knowledge graphs (KGs)~\cite{chen2024knowledge,shengyuan2024differentiable}. It has various applications, especially for assisting the visually impaired users~\cite{gurari2018vizwiz}, yet still, a challenging task that requires complex reasoning across different data modalities~\cite{yu2020cross}.

\begin{figure}[!t]
    \centering
    \includegraphics[width=7.5cm]{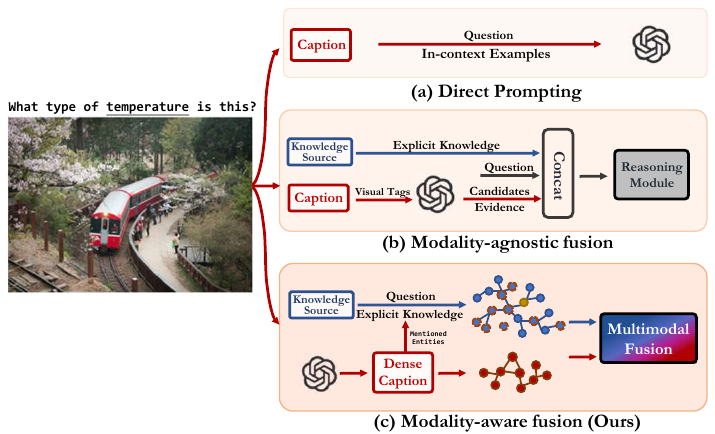}
    \captionsetup{font={small}}
    \caption{A sketched comparison on employing LLMs for KVQA between existing learning paradigms and ours. 
    }
    \label{fig:running}
    \vspace{-5mm}
\end{figure}

Recently, several studies have explored using large language models (LLMs) as supplementary knowledge bases and reasoning tools for KVQA~\cite{pica,gui2022kat,lin2022revive}; according to how they fuse the knowledge, they can be broadly categorized into direct prompting and modality-agnostic approaches, shown in Figure~\ref{fig:running} (a) and (b), respectively. The former directly prompts the question and the corresponding image caption to LLMs for answers~\cite{pica}. The latter leverages LLMs to generate candidate answers with supporting evidence and simply combines both question and the external knowledge embedding, e.g., Wikidata~\cite{vrandevcic2014wikidata}, for reasoning at the final stage~\cite{gui2022kat,lin2022revive}.

While the above methods have employed LLMs in various ways for KVQA, we argue that they have not fully leveraged the knowledge from LLMs and lack the cross-modal reasoning ability, potentially resulting in sub-optimal performance for complex VQA scenarios. \((i)\) LLMs could incorrectly answer questions or provide unreliable evidence for reasoning. On the one hand, direct prompting to LLMs may struggle to identify the right answer for many complex or domain-specific questions, due to the lack of domain knowledge~\cite{amaro2023ai,shen2023chatgpt}. On the other hand, LLMs may be prone to generating hallucination~\cite{gravel2023fabricate,bang2023multitask} and producing misleading evidence in support of candidate answers. \((ii)\) Integrating multimodal knowledge in a modality-agnostic manner can be sub-optimal. Specifically, existing methods simply concatenate different modal representations, e.g., questions, captions, tags, and external knowledge, for reasoning. This design lacks the necessary cross-modal exchange to enrich the semantics of entities, limiting the final reasoning performance. For example, to correctly answer the question in Figure~\ref{fig:running}, the model is required to infer the season based on a cross-modal understanding of the inputs, such as the ``keep warm'' purpose of ``coat'' and the ``spring blooming'' feature of ``sakura''.

In this work, we study the following research question: \emph{\textbf{How can we effectively leverage the knowledge from LLMs to enhance the comprehensive understanding and reasoning of the images and questions in KVQA?}} Answering this question is nontrivial due to the following challenges. \((i)\) It is hard to properly incorporate the knowledge from LLMs. LLMs may generate hallucinations when dealing with requests that are not covered in their training corpus. Simply prompting them may generate noisy and irrelevant responses. \((ii)\) Semantic alignment of multiple knowledge sources is challenging. Given image captions, object/region features, external knowledge from KGs, and implicit knowledge from LLMs, appropriately aligning relevant semantic information in different modalities cannot be readily achieved.

To tackle these challenges, we present a novel modality-aware framework to effectively integrate LLMs for KVQA in Figure~\ref{fig:running} (c), dubbed \texttt{MAIL}. Specifically, \((i)\) we propose a two-stage prompting strategy to maximally leverage the knowledge from LLMs for image understanding. We initialize a dense caption by prompting a visual LLM, e.g., Visual ChatGPT~\cite{wu2023visual} and MiniGPT-4~\cite{zhu2023minigpt}. To depict the detailed visual scenes in the caption, we construct a \textit{scene graph} by defining twelve condensed relations and prompting the LLM to extract spatial and object features accordingly in the form of triples, e.g., (\textit{sakura}, \textit{at\_location}, \textit{tree}). \((ii)\) We integrate the external knowledge from KGs to form a coupled \textit{concept graph}, where the mentioned entities in scene graphs are linked with real-world assertions and facts to facilitate knowledgeable reasoning, such as (\textit{coat}, \textit{used\_for}, \textit{keep warm}) and (\textit{sakura}, \textit{typle\_of}, \textit{spring blooming}). \((iii)\) A tailored pseudo-siamese graph medium fusion is designed for effective multimodal graph fusion. Inspired by the success of pseudo-siamese network in measuring the similarity of two correlative inputs~\cite{xia2021pseudo,gupta2023siamese}, we extend it to graphs to process intra-modal information. It consists of two graph attention networks with the same architecture but different weights. In each sub-encoder, we concentrate on one modality and design a tailored context-aware propagation. This guides our model to attentively prioritize the most valuable entities subject to the particular question. Then we leverage the shared mentioned entities in both coupled graphs as mediums to bridge the cross-modal interaction. The model continuously exchanges their embeddings between two modalities, bringing sufficient complementary knowledge to the other modality respectively. It merely allows inter-modal exchanging by constraining it within the mediums. In general, \texttt{MAIL} effectively enhances a tight inter-modal fusion while maximally preserving the insightful intra-modal information for each modality. 

Our major contributions are summarized below:
\begin{itemize} [leftmargin=*]
\item We formally define a novel learning paradigm, modality-aware integration with LLMs for knowledge-based visual question answering.
\vspace{-2mm}
\item The implicit knowledge in LLMs is carefully leveraged via an effective prompting strategy for coupled scene/concept graph construction.
\vspace{-2mm}
\item We further propose a tailored pseudo-siamese graph medium fusion to integrate multimodal knowledge sources. It balances both intra-modal processing and inter-modal exchange.
\vspace{-2mm}
\item Extensive experiments are conducted on two benchmark datasets. \texttt{MAIL} significantly achieves superior performance over a variety of state-of-the-art baselines with 24$\times$ less computational resources and $2\sim4\times$ faster inferential time.
\end{itemize}

\section{Problem Statement}
KVQA requires the model to provide answers to the question $\mathcal{Q}$ of the corresponding image $\mathcal{I}$ based on external knowledge $\mathcal{G}$. In this paper, we propose a novel learning paradigm for leveraging LLMs $f(\cdot)$ for comprehensive knowledge-based VQA.
\fbox{\parbox{0.465\textwidth}{ \vspace{0.05cm}
Given an image $\mathcal{I}$, a relevant question $\mathcal{Q}$ and external knowledge $\mathcal{G}$, we aim to integrate a visual LLM $f(\cdot)$ and fuse $\left\{f(\mathcal{I}),\mathcal{Q},\mathcal{G}\right\}$ for prediction. The overall performance is evaluated by the accuracy of returned answers with the ground truths.
}}

\begin{figure*}[!t]
\centering
    \includegraphics[width=16cm]{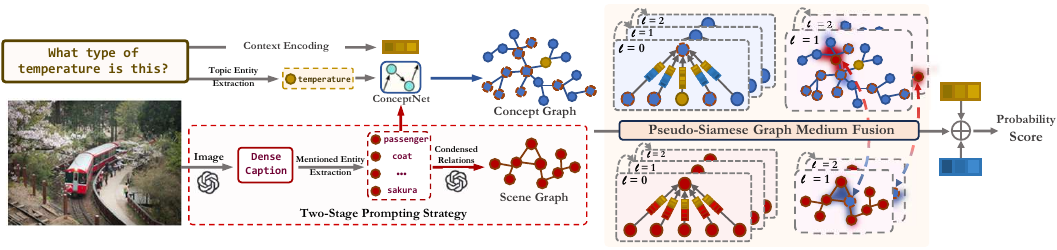}
    \caption{Our proposed framework \texttt{MAIL}, a novel modality-aware integration for knowledge-based VQA with LLMs. Nodes in \textbf{\textit{\textcolor[RGB]{0,0,205}{blue}}} stand for external knowledge, while \textbf{\textit{\textcolor{purple}{red}}} is for visual objects and \textbf{\textit{\textcolor[RGB]{205,155,29}{yellow}}} shows the topic entities from questions. Blue nodes with red dashed borders indicate the extracted mediums in concept graph.  \texttt{MAIL} is trained to integrate multimodal information for comprehensive cross-modal reasoning with a tailored PS-GMF.} 
    \label{fig:fig1}
\end{figure*}

\section{Methodology}
In this section, we introduce the detailed rationale of our proposed framework. An illustration of \texttt{MAIL} is shown in Figure~\ref{fig:fig1}. 
We first carefully leverage the knowledge from LLMs for coupled graph construction. Then, we formulate the pseudo-siamese graph medium fusion (PS-GMF). Through an effective integration of two tailored training objectives, we jointly optimize the model for accurate prediction.

\subsection{Scene Graph Construction}
\textbf{Dense Caption Generation}
We carefully design a hard prompt that requires a visual LLM $f(\cdot)$ to depict the detailed appearance of all the objects in the image and the spatial relations between them. We obtain the generated caption through
\vspace{-2mm}
\begin{equation}
    \mathcal{D} = f(\mathcal{I},Prompt). 
    \vspace{-2mm}
\end{equation}
We consider the identified visual entities in the image as key mentioned entities appearing in the caption, denoted as $\mathcal{M}=[m_{1}, m_{2}...m_{n}] \in \mathcal{D}$. They significantly dominate the multimodal information of both visual features and external knowledge required to answer the questions.\\
{\bf Prompt-enhanced Triple Extraction}\\
Given the extracted mentioned entities, we employ LLMs to extract triples. To fully leverage LLMs' comprehension of image captions and prioritize the important visual features, we pre-define 12 relations $\mathcal{R}=[r_1,r_2,...r_{12}]$ from two aspects: \((i)\) Spatial features. We constrain the description with \textit{at\_location}, \textit{next\_to}, \textit{in\_front\_of}, \textit{surrounded\_by}, \textit{covered\_by}, \textit{includes} and \textit{holds}. \((ii)\) Object features are preserved with not only visual outlooks, i.e., \textit{has\_property}, \textit{has\_color}, \textit{made\_of} and \textit{wears}, but also the intentions of the object if he/she is a human, i.e., \textit{intends\_to}. We design a hard template to prompt LLMs for scene graph construction as, 
    \vspace{-2mm}
\begin{equation}
    \mathcal{G}^{S} = f(Prompt, \mathcal{D}, \mathcal{M}, \mathcal{R} ).
    \vspace{-2mm}
\end{equation}
We show the detailed statistics and beautiful distributions of all twelve condensed relations in both benchmarks OK-VQA~\cite{marino2019ok} and FVQA~\cite{wang2017fvqa} in \textbf{Appendix Table~\ref{relations_ok}}.

\subsection{Concept Graph Construction}
In parallel, we incorporate ConceptNet~\cite{speer2018conceptnet} for external commonsense knowledge to construct a concept graph. It is one of the largest knowledge graphs that provides a myriad of structured triples and contains more than eight million real-world entities. We link each mentioned entity $m$ and the topic entity in the question with ConceptNet, and denote the constructed graph as $\mathcal{G}^{C}$ with sufficient textual descriptions, attributes, categories, and properties of $\mathcal{M}$, that are not present in the image so as to facilitate a more knowledgeable reasoning background for various questions.

\subsection{Pseudo-siamese Graph Medium Fusion}
Typical pseudo-siamese networks (PSNs) could effectively measure the similarity between two inputs~\cite{gupta2023siamese,xia2021pseudo}. We extend it to graphs, which naturally fit the requirement of learning coupled graphs for intra-modal processing, leading to pseudo-siamese graph neural networks (PSGs). However, PSG is incapable of cross-modal fusion. Particularly equipped for PSG to enable inter-modal learning, we further design a \textit{graph medium fusion} (GMF) algorithm.\\
{\bf Pseudo-siamese Graph Neural Network}\\
Locating valuable entities in different modalities is essential for KVQA. Here, we instantiate PSG with a novel context-aware message propagation scheme to prioritize the most important knowledge in each modality subject to the question context. \\
\textit{\textbf{Definition.}} [\textbf{Pseudo-siamese GNN}]
\textit{We refer to a pseudo-siamese graph neural network that consists of two identical graph neural networks for two relevant inputs. They share the same architecture, i.e., attention mechanism, aggregation function, combination function and activation function, but different weights.}

\renewcommand{\arraystretch}{1.2}
\begin{table}[t!] \footnotesize
    \centering
    \setlength{\tabcolsep}{2mm}
    \begin{tabular}{c|c}
    \hline
    \textbf{PSG Architecture} & \textbf{Formulated Definition}  
    \cr \cmidrule{1-2}
    Context-aware Attention & $\boldsymbol{\Phi}\left(\boldsymbol{m}_{t} \| \mathbf{c}\right)$ \\
    Aggregation Function & $\sum_{t\in \mathcal{N}_{h}} \alpha_{\boldsymbol{m}_{t}} \times \boldsymbol{m}_{t}$\\
    Combination Function & $J(\boldsymbol{e}_{\mathcal{N}_{h}}^{\ell})$ + $\boldsymbol{e}_{h}^{\ell}$ \\
    Activation Function & $\begin{cases}
        \boldsymbol{e}_{h}, & \textit{if } \boldsymbol{e}_{h} \geq 0, \\
        (1e-2)\times\boldsymbol{e}_{h}, & \textit{otherwise}.
    \end{cases}$\\
    \hline
    \end{tabular}  
    \caption{Formulated definitions of the shared architectures for two sub-networks in the proposed Pseudo-Siamese Graph Neural Network.}
    \label{Architecture}
    \vspace{-5mm}
\end{table}
As two sub-networks in PSG share the same architecture, we uniformly provide formulations for the intra-modality processing. For each head entity $h$, we aggregate all the messages from its neighbor tail entities, this set of neighbors is denoted as $\mathcal{N}_h$ and $t\in\mathcal{N}_h$. Since relations in multimodal graphs contain indispensable information for reasoning various real-world questions, we establish the message passing at the triple level, i.e., $(h,r,t)$ to capture abundant semantics as follows.
\begin{equation}
   \boldsymbol{m}_{t\in\mathcal{N}_h} = \boldsymbol{W}(\boldsymbol{e}_{h},\boldsymbol{e}_{r},\boldsymbol{e}_{t}),
\end{equation}
where $(\boldsymbol{e}_{h},\boldsymbol{e}_{r},\boldsymbol{e}_{t})$ is the triple embedding associated with $(h,r,t)$, and $\boldsymbol{W}$ is a learnable matrix for linear transformation. We initialize the entity and relation embedding with a pre-trained language model RoBERTa-large~\cite{liu2019roberta}.

While multimodal graphs always contain desperate information with each other, uniformly training each subnetwork in PSG based on the final prediction lacks awareness of the multimodal characteristics. To this end, we design tailored graph attention networks~\cite{han2022g,NEGCN} that allocate a \textit{context-aware} weight $\hat{a}$ to each message~\cite{dong2023hierarchy}, only prioritizing the multimodal messages in both coupled graphs that are highly related to the question. $\hat{a}_{\boldsymbol{m}_{t}}$ for each message $\boldsymbol{m}_{t}$ is correspondingly computed as:
\vspace{-3mm}

\begin{equation}
    \hat{a}_{(h,r,t)} = \boldsymbol{\Phi}\left(\boldsymbol{m}_{t} \| \mathbf{c}\right),
\end{equation}
where $\boldsymbol{\Phi}$ is the adopted activation function, i.e., LeakyReLU. We endow the attention mechanism to be context-aware by concatenating the question context embedding $\boldsymbol{c}$, expressed as $\|$. Notably, we fix the question context embedding $\boldsymbol{c}$ with RoBERTa and only allow it to participate during the attention allocation process.

By normalizing the attention scores obtained previously, we further assign normative values $\alpha$ to each message $\boldsymbol{m}_t$ of $(h,r,t)$:
\vspace{-2mm}

\begin{equation}\small
    \alpha_{\boldsymbol{m}_{t}} = \frac{\hat{a}_{(h,r,t)}} {\sum_{(h,r',t')\in \mathcal{N}_h}\hat{a}_{(h,r',t')}}.
\end{equation}

To this end, with a weighted sum aggregation operator, we are able to acquire the aggregated representation for entity $h$ in the current layer from its neighbors as $   \boldsymbol{e}_{\mathcal{N}_{h}}^{\ell} =\sum_{(h,r,t)\in \mathcal{N}_{h}} \alpha_{(h,r,t))} \times \boldsymbol{m}_{t}^{\ell}$, where the layer number in PSG is denoted as $\ell$. We summarize the major functions in Table~\ref{Architecture}. We finalize the overall architecture of PSG for both inputs from scene graph $\mathcal{G}^{S}$ and concept graph $\mathcal{G}^{C}$.

\begin{equation}\small
\begin{aligned}
    \boldsymbol{e}_{h}^{S(\ell+1)}  &=  J( \sum_{(h,r,t)\in \mathcal{N}_{h}} \alpha_{(h,r,t))} \times \boldsymbol{m}_{t} ) + \boldsymbol{e}_{h}^{S(\ell)},\\
    \boldsymbol{e}_{\hat{h}}^{C(\ell+1)}  &=  J( \sum_{(\hat{h},\hat{r},\hat{t})\in \mathcal{N}_{\hat{h}}} \alpha_{(\hat{h},\hat{r},\hat{t})} \times \boldsymbol{m}_{\hat{t}} ) + \boldsymbol{e}_{\hat{h}}^{C(\ell)},
    \label{equation: CGarchitecture}
\end{aligned}
\end{equation}
where $J$ is a multi-layer perception. The model effectively combines the learned neighbor information $\boldsymbol{e}_{\mathcal{N}_{h}}^{\ell}$ and itself $\boldsymbol{e}_{h}^{\ell}$ in current layer. We obtain final representations of all the entities when the layer number $\ell$ reaches the pre-defined target.\\
{\bf Graph Medium Fusion}\\
In this subsection, we aim to fill the gaps of the aforementioned PSG on \textit{inter-modal} learning. However, there is a challenging dilemma centered around striking the right balance between two crucial aspects. On one hand, we want to maximize the \textit{inter-modal} fusion, where multimodal information could collaborate to yield a more insightful and nuanced understanding of the underlying knowledge subject to answering the question. On the other hand, we recognize the necessity of preserving the integrity of \textit{intra-modal} processing. Considering excessive inter-modal fusion could introduce noise from each other, we aim to maintain the distinctive characteristics and valuable insights that each modality inherently holds. 

Since the mentioned entities $\mathcal{M}=[m_{1}, m_{2}...m_{n}]$ are shared by $\mathcal{G}^{S}$ and $\mathcal{G}^{C}$, we consider these entities existing in both coupled graphs as mediums that possess similar embeddings, since they represent the same real-world object though appearing in different modalities. Motivated by this, we design a novel graph medium fusion algorithm that leverages the medium to bridge two modalities. To get rid of the dilemma, we \((i)\) exchange the representations of mediums $\boldsymbol{e}_m$ within their respective graphs. This allows the model to delicately introduce cross-modal information with their neighbor entities in the respective graphs, i.e., $\boldsymbol{e}_{\mathcal{N}_{m}}$; \((ii)\) We strictly impose restrictions on the cross-modal exchange to be within the mediums. This gently brings two modalities closer to each other, while maximally maintaining their individualities. The formulated graph medium fusion process between the coupled graphs is written below.
\vspace{-2mm}
\begin{equation}\small
\begin{aligned}
    e^{S}_{m}=
    \begin{cases}
        \boldsymbol{e}^{S}_{m}, & \textit{if } \ell = 0, \\
        \boldsymbol{e}^{C}_{m}, & \textit{otherwise}.
    \end{cases}
    \end{aligned}
    \quad
    \begin{aligned}
    e^{C}_{m}=
    \begin{cases}
        \boldsymbol{e}^{C}_{m}, & \textit{if } \ell = 0, \\
        \boldsymbol{e}^{S}_{m}, & \textit{otherwise}.
    \end{cases} 
    \vspace{-3.5mm}
\end{aligned}
\end{equation} 

Specifically, we froze the medium embeddings in the first layer to ensure they have initially aggregated important 1-hop neighbor information. Afterward, the embeddings for the same medium are automatically exchanged after message-passing in the current layer. This sequential approach ensures a high-quality exchange of information between modalities, i.e., visual features and external knowledge, while initially preserving the local context within each modality before they engage in cross-modal interactions during the following layers.

\subsection{Training Objective}
{\bf Answer-targeted Inferential Loss}\\
The primary target of our model is to accurately predict the final answer subject to the particular image and question context. We adopt the binary cross-entropy loss to optimize the inferential performance:
\vspace{-3mm}
\begin{equation}\small
    \vspace{-2mm}
    \mathcal{L}_{Inference} = -log\frac{MLP(\boldsymbol{e}_{a} + \mathbf{c})} {\sum_{a'\in {\mathcal{G}^{C}}}MLP(\boldsymbol{e}_{a'} + \mathbf{c})},
    \vspace{1mm}
\end{equation}
where $a$ is the correct answer and $a'$ is one of all the candidate answers from $\mathcal{G}^{C}$. We employ $MLP(\boldsymbol{e}_{a}+\boldsymbol{c})$ to compute the probability of all the candidate entities in $\mathcal{G}^{C}$ and prioritize the highest one as the final answer.\\
{\bf Maximum Mean Discrepancy loss}\\
Based on the assumption that one medium in two modalities should be similar to the maximum extent, we approximate their similarity by adopting an auxiliary loss, i.e., Maximum Mean Discrepancy (MMD) loss. The basic kernel function is formulated as follows:
\begin{equation}\small
    \vspace{-2mm}
    \mathcal{K}(\boldsymbol{e}_{m}^{S}, \boldsymbol{e}_{m}^{C}) = \exp\left(-\frac{||\boldsymbol{e}_{m}^{S} - \boldsymbol{e}_{m}^{C}||^2}{2 \sigma^2}\right),
\end{equation}
where $\mathcal{K}$ represents the kernel function and $\sigma$ is a hyperparameter controlling the width of the kernel~\cite{steinwart2012mercer}. Given a valid kernel function where $\mathcal{K}(\boldsymbol{e}_{m}^{S},\boldsymbol{e}_{m}^{C}) = (\phi(\boldsymbol{e}_m^{S})-\phi(\boldsymbol{e}_m^{C}))$, we denote the corresponding feature mapping function as $\phi$. The final MMD loss for cross-modal alignment is demonstrated hereunder,
\vspace{-5mm}

\begin{equation}\small
    \vspace{-2mm}
    \mathcal{L}_{Medium} = ||\frac{1}{n} \sum_{m\in \mathcal{M}} \phi(\boldsymbol{e}_m^{S}) - \frac{1}{n} \sum_{m\in \mathcal{M}}\phi(\boldsymbol{e}_m^{C}) ||^2.
\end{equation}
We aim to minimize this loss to encourage the learned representations for the same medium from two modalities to be similar in the shared PSG architecture. This effectively guides the process of graph medium fusion by constraining the similarity of mediums in different modalities with each other.

\subsubsection{Joint Optimization}
The overall framework is jointly optimized according to training objectives as aforementioned. Despite the effectiveness of $\mathcal{L}_{Medium}$, it may introduce inevitable noise by irrespectively forcing the mediums from two modalities to be exactly aligned, which ignores the nature of different modalities. To alleviate this problem, we introduce a hyperparameter $\lambda$ to control the contribution from $\mathcal{L}_{Medium}$. To this end, the final training loss is calculated below:
\vspace{-6mm}

\begin{equation}
\vspace{-1mm}
    \mathcal{L}_{Joint} = \mathcal{L}_{Inference} + \lambda \mathcal{L}_{Medium}.
    \vspace{-2mm}
\end{equation}

\renewcommand{\arraystretch}{1}
\begin{table*}[t!] \footnotesize
    \vspace{-3mm}
    \centering
    \setlength{\tabcolsep}{0.5mm}
    \begin{tabular}{c|c|c|c|c}
    \hline
    \textbf{Method} & \textbf{Model Inputs} & \textbf{External Knowledge} & \textbf{Fusion Strategy} & \textbf{Acc. (\%)}
    \cr \cmidrule{1-5}
    Q Only &Question + Image &- &- & 14.93
    \cr \cmidrule{1-5}\hline  
    \multicolumn{5}{c}{\textbf{Traditional End-to-end Baselines}}
    \cr \cmidrule{1-5}
    BAN &Question + Image &- & -  & 25.17\\
    BAN +AN &Question + Image &Wikipedia & Modality-agnostic  & 25.61\\
    MUTAN &Question + Image &- & - & 26.41\\
    MUTAN +AN &Question + Image &Wikipedia & Modality-agnostic  & 27.84\\
    ConceptBERT &Question + Image & ConceptNet & Modality-agnostic & 33.66\\
    HCNMN &Question + Image  & WordNet & Modality-agnostic &36.74\\
    Krisp &Question + Image & Wikipedia + ConceptNet & Modality-agnostic & 38.90\\
    MAVEx &Question + Image & Wikipedia + ConceptNet + Google Images & Modality-agnostic & 41.37 \\
    VLC-BERT &Question + Image & COMET + ConceptNet & Modality-agnostic & 43.14\\
    MCAN &Question + Image & - & - & 44.65 
    \cr \cmidrule{1-5}\hline
    \multicolumn{5}{c}{\textbf{Large Language Model-enhanced Baselines}} 
    \cr \cmidrule{1-5}
    PICa-Base &Question + Caption + Object Tags & Frozen GPT-3 (175B) & - & 43.30\\
    Pica-Full &Question + Caption + Object Tags & Frozen GPT-3 (175B) & - & 48.00
    \cr \cmidrule{1-5}
    KAT (Single) &Question + Caption + Object Tags & Frozen GPT-3 (175B) + Wikidata & Modality-agnostic & 53.09\\
    KAT (Ensemble) &Question + Caption + Object Tags & Frozen GPT-3 (175B) + Wikidata & Modality-agnostic & 54.41\\
    REVIVE  &Question + Caption + Region Tags & Frozen GPT-3 (175B) + Wikidata & Modality-agnostic & 53.83
    \cr \cmidrule{1-5}

    \texttt{MAIL} (ours) & Question + Image & Frozen MiniGPT-4 (7B)$^*$ + ConceptNet &Modality-aware & \textbf{56.69}
    \cr \cmidrule{1-5}\hline
    \end{tabular}   \\
    \caption{The overall performance comparison on benchmark dataset OK-VQA. We also elaborate on the detailed comparison with a variety of baselines on the knowledge sources that support their inference, i.e., model inputs, external knowledge, as well as how they fuse multiple modalities.}
    \footnotesize{$^*$ We merely leverage it for caption and scene graph construction, with no extra information that is not present in the images.}
    \label{main_results}
    \vspace{-1mm}
\end{table*}

\section{Experiments}
In this section, we conduct a variety of experiments to demonstrate the effectiveness of our proposed \texttt{MAIL}. We aim to answer four research questions:
\begin{itemize}[leftmargin=*]
    \item \textbf{RQ1 (Main Results):} How does \texttt{MAIL} perform compared with different types of SOTA models?\vspace{-2mm}
    \item \textbf{RQ2 (Hyperparameter analysis):} How do hyperparameters influence the performance?\vspace{-2mm}
    \item \textbf{RQ3 (Ablation studies):} Does each component eventually contribute to the overall performance?\vspace{-2mm}
    \item \textbf{RQ4 (Case study):} How effectively does \texttt{MAIL} work in real-world VQA tasks?
\end{itemize}
\subsection{Experimental Setup}
{\bf Datasets}\\
Following the previous work~\cite{marino2021krisp,pica,gui2022kat,mavex,lin2022revive}, we mainly conduct our experiments on \textbf{OK-VQA}~\cite{marino2019ok}, which is currently the largest and most challenging benchmark, consisting of 14,055 image-question pairs. To further demonstrate the generalization, we also experimentalize on \textbf{FVQA} dataset~\cite{wang2017fvqa}, which was the first exploration of KVQA.\\
\renewcommand{\arraystretch}{1}
\begin{table}[!ht] \footnotesize
    \centering
    \setlength{\tabcolsep}{4.4mm}
    \begin{tabular}{c|c|c}
    \hline\hline
    \textbf{Method} & \textbf{Fusion Strategy} & \textbf{Acc. (\%)}
    \cr \cmidrule{1-3}
    XNM  & Modality-agnostic & 63.74\\
    KI-Net  & Modality-agnostic  & 63.78\\
    UnifER  & Modality-agnostic & 66.83\\
    MCAN  & - & 64.47\\
    HCNMN & Modality-agnostic &69.43\
    \cr \cmidrule{1-3}
    \texttt{MAIL} (ours) &Modality-aware & \textbf{73.95}
    \cr \cmidrule{1-3}\hline
    \end{tabular}   \\
    \caption{Performance comparison on FVQA.}
    \label{main_results_f}
    \vspace{-5mm}
\end{table}
{\bf Baselines}\\
We adopt two pipelines of off-the-shelf methods for performance comparison. Details are demonstrated in the \textbf{Appendix~\ref{basel}}. \((i)\) Traditional end-to-end baselines that design various multimodal learning algorithms for final reasoning over the posed questions. \((ii)\) LLM-enhanced baselines that leverage LLMs, i.e., GPT-3, for direct answer prediction or relevant supporting evidence generation.

\vspace{-2mm}
\subsection{Main Results}
To answer \textbf{RQ1}, in Table~\ref{main_results} \&~\ref{main_results_f}, we summarize the comparisons with all the important baselines. The performance is evaluated by the soft accuracy following previous research~\cite{promptcap}. \texttt{MAIL} outperforms all the traditional baselines regardless of their various knowledge sources and the advantages of leveraging a feature-level image representation. 
\texttt{MAIL} achieves 12.04\% improvements over the best traditional baseline, i.e., MCAN, on OK-VQA and 14.7\% on FVQA. For LLM-enhanced baselines, it is worth mentioning that they have utilized the generative ability from~\cite{lin2022revive,gpt3}, which makes them especially advantageous in answering subjective questions, for instance, `Can people travel on the freeway' or `Is it illegal?'. Despite this, \texttt{MAIL} still outperforms the best LLM-enhanced baseline with 2.28\% increases in general, let alone 13.39\% over PICa.

\begin{table}[t]\footnotesize
\centering
\setlength{\tabcolsep}{2.5mm}
\begin{tabular}{cccccc}
\toprule
  \textbf{ACC.\% }   & $\ell = 2$ & $\ell = 3$ &$\ell = 4$  &$\ell = 5$   &$\ell = 6$ \\ \midrule
\texttt{MAIL} & 56.41 & \textbf{56.69} & 55.45    & 54.11  & 52.80     \\ \bottomrule
\vspace{-3mm}
\end{tabular}
\caption{Evaluation on the influences of graph layers in pseudo-siamese graph medium fusion.}
\label{graph layer}
\vspace{-4mm}
\end{table}

\begin{table}[t]\footnotesize
\centering
\setlength{\tabcolsep}{1mm}
\begin{tabular}{ccccccc}
\toprule
  \textbf{ACC.\% }   & $\lambda = 0$ & $1e-5$ &$ 1e-4$  &$1e-3$ &$ 1e-2$   &$1e-1$ \\ \midrule
\texttt{MAIL} & 53.34 & 54.18 & 55.31  & \textbf{56.69}  & 54.30 & 55.82   \\ \bottomrule
\vspace{-3mm}
\end{tabular}
\caption{Exploring the control over the impacts from $\mathcal{L}_{Medium}$ to preserve insightful intra-modal learning.}
\label{lambda}
\vspace{-4mm}
\end{table}

\begin{figure*}[!t]
\centering
    \includegraphics[width=15.8cm]{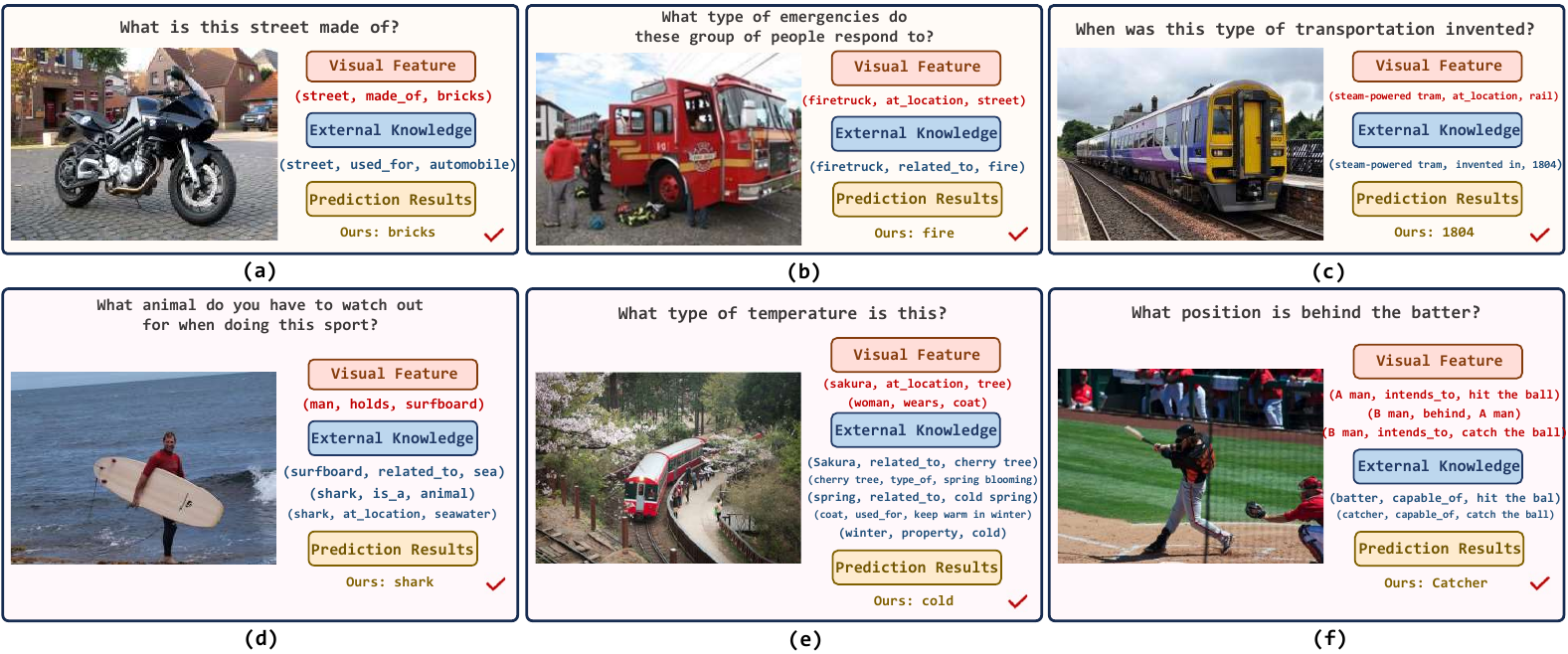}
    \caption{Case studies with both single-hop and multi-hop reasoning examples in OK-VQA.} 
    \label{fig:case}
    \vspace{-3mm}
\end{figure*}

Moreover, \texttt{MAIL} is resource-efficient, requiring the smallest number of parameters among all the LLM-enhanced baselines, shown in Table~\ref{paramsize}. We have used significantly far fewer parameters than any other LLM-enhanced models, i.e., 7.13 B, for answer prediction. As a result, the inferential time of \texttt{MAIL} for one test question is 0.661s (when batch size = 1).
Generally, existing LLM-enhanced baselines commonly utilize over 24 times more parameters and 2$\sim$4 times of inferential time than \texttt{MAIL}.

\renewcommand{\arraystretch}{1.2}
\begin{table}[t!] \footnotesize

    \centering
    \setlength{\tabcolsep}{0.5mm}
    \begin{tabular}{c|c|c|c}
    \hline\hline
    \textbf{Models} & \textbf{$\sim$Param Size} & \textbf{Training Time} & \textbf{Inference Time} 
    \cr \cmidrule{1-4}
    PICa & $\sim$175.00 B & / & 1.547 s\\
    KAT & $\sim$175.80 B & 3.025 s & 1.292 s \\
    REVIVE & $\sim$175.80 B & 4.500 s & 2.644 s\\
    \texttt{MAIL}(Ours) & \textbf{$\sim$7.13 B} & \textbf{2.699 s} & \textbf{0.661 s} 
    \cr \cmidrule{1-4}\hline
    \end{tabular}
    \caption{Comparisons on the computational costs and inferential time with LLM-enhanced baselines.}
    \label{paramsize}
    \vspace{-4mm}
\end{table}

\renewcommand{\arraystretch}{1.2}
\begin{table}[!t] \footnotesize
    \centering
    \setlength{\tabcolsep}{6mm}
    \begin{tabular}{c|c}
    \hline\hline
    \textbf{Reasoning Module} & \textbf{Accuracy (\%)}  
    \cr \cmidrule{1-2}
    PSG (w/o GMF) & 55.53 \\
    PS-GMF & \textbf{56.69} 
    \cr \cmidrule{1-2}\hline
    \end{tabular}  
    \caption{Verification of the importance of inter-modality fusion by removing GMF with PSG only.}
    \label{psg}
    \vspace{-5mm}
\end{table}

\subsection{Hyperparameter Analysis}
{\bf Search of Graph layers}\\
The main architecture of PS-GMF naturally comprises the discussion of the impacts from graph layers $\ell$. We empirically hypothesize that augmenting the depth of the $\ell$ could facilitate both a deeper understanding of single modalities (i.e., PSG) and a more profound exchange of information between modalities (i.e., GMF). However, it remains unclear about when to reach the plateau. Simply adding more layers may over-fuse two modalities and lose the ability of intra-modal processing, while reducing layers may lead to an adverse situation with inadequate inter-modal fusion. To this end, we vary the layer number and show the performance changes in Table~\ref{graph layer}. The final prediction performance of \texttt{MAIL} is reported when $\ell = 3$.\\
{\bf Investigation on hyperparameter $\lambda$}\\
While an excessively strict alignment of mediums may homogenize the
intra-modal information, we aim to find a suitable $\lambda$ that constrains the impacts of $\mathcal{L}_{Medium}$. This could significantly encourage harmonious inter-modal fusion from multiple modalities while retaining the richness and specificity inherent to each modality. The experimentation process involves a systematic adjustment of $\lambda$ across a range of values, specifically within the interval $[0, 1e-5, 1e-4, 1e-3,1e-2,1e-1]$. We showcase the results in Table~\ref{lambda}. Upon careful examination of the performance trends, we employ $\lambda= 1e-3$ for a balanced trade-off.

\renewcommand{\arraystretch}{1.2}
\begin{table}[!t] \footnotesize

    \centering
    \setlength{\tabcolsep}{1mm}
    \begin{tabular}{c|c|c}
    \hline
    \textbf{Pure LLMs} & \textbf{Multimodal Understanding} & \textbf{Acc.(\%)}  
    \cr \cmidrule{1-3}\hline  
    \multicolumn{3}{c}{\textbf{Large Language Models}}
    \cr \cmidrule{1-3}
    Llama (7B) & Dense Caption & 39.27 \\
    Llama2 (7B) & Dense Caption & 45.35\\
    ChatGPT (GPT3.5) & Dense Caption & 40.26\\
    GPT-4 & Dense Caption & 54.33
    \cr \cmidrule{1-3}\hline     
    \multicolumn{3}{c}{\textbf{Visual Large Language Models}}
    \cr \cmidrule{1-3}
    Visual ChatGPT & BLIP-VQA-Base + GPT3.5  & 38.70\\
    MiniGPT-4 (7B) & ViT + Vicuna & 51.26     
    \cr \cmidrule{1-3}\hline
    Ours & Dense Caption + PS-GMF & \textbf{56.69}\\
    \hline
    \end{tabular}   
    \caption{Ablation studies on comparing with pure LLMs by directly feeding the questions and \((i)\) corresponding image caption to LLMs or \((ii)\) the raw images to visual LLMs for answers in a zero-shot setting.}
    \label{ablation_LLM}
    \vspace{-5mm}
\end{table}

\subsection{Ablation Studies} 
{\bf Empirical comparison with LLMs}\\
In this ablation study, we further demonstrate our tailored multimodal learning module PS-GMF, and delineate the specific contributions by comparing it against frozen LLMs. Specifically, we adopt both pure LLMs, i.e., Llama~\cite{touvron2023llama} and Llama2~\cite{touvron2023llama2}, as well as visual LLMs with Visual ChatGPT~\cite{wu2023visual} and MiniGPT-4~\cite{zhu2023minigpt}. We exclusively constrain the inputs in a zero-shot setting with only dense captions and questions for LLMs, while raw images and questions for frozen visual LLMs. The results are summarized in Table~\ref{ablation_LLM}. \texttt{MAIL} outperforms the best LLM GPT-4 with 2.36\% improvements, attributed to the effective graph medium fusion that integrates external knowledge. \texttt{MAIL} also significantly outperforms Visual ChatGPT and MiniGPT-4 with 17.99\% and 5.43\% higher accuracy. The results shed light on the cross-modal reasoning ability of \texttt{MAIL}. \\
{\bf Reasoning with PSG Only}\\
In this subsection, we explore the importance of inter-modality interaction by removing the graph medium fusion and only relying on PSG for inference. We list the performance of `PSG w/o (GMF)' in Table~\ref{psg}. The complete multimodal reasoning with PS-GMF outperforms the version with only intra-modal learning with 1.16\% improvements. Under this PSG-only setting, we seek to grasp insights into the necessity of graph medium fusion for fostering effective inter-modality interaction. Understanding the performance impact of omitting this fusion mechanism supports the value of shared entities and medium exchange in bridging the cross-modal interaction and facilitates our proposed modality-aware integration with LLMs.

\subsection{Case Studies}
In this section, we answer \textbf{RQ4} with six real-world examples from OK-VQA in Figure~\ref{fig:case} to shed light on our effectiveness. \textbf{Single-hop questions} can be directly inferred with easily accessible information from either the visual content or external knowledge sources, while \textbf{multi-hop questions} pose more challenges for accurately locating answers several hops away from mentioned entities.

These cases show the adeptness of \texttt{MAIL} in handling a spectrum of questions, requiring both straightforward inferences from explicit information and complex multi-hop reasoning ability by integrating implicit knowledge sources. For example, Figure~\ref{fig:case} (a) can be answered based on the visual information captured by the scene graph without external knowledge, while the answer of (e) needs to be artfully inferred from two different angles, i.e., both the blossom season of sakura and the warmth of people's clothes. These can be attributed to \((i)\) the coupled graph construction that contains abundant modality-aware knowledge to ground the reasoning, as well as \((ii)\) the effective design of our pseudo-siamese graph neural network. It benefits sufficient preservation of intra-modal information and adequate cross-modal fusion, resulting in a powerful multi-hop reasoning ability over both inherent visual features and external knowledge.
\vspace{-2mm}

\section{Related Work}
\textbf{KVQA with KGs.} Early studies either dedicated to integrating different knowledge sources~\cite{wang2017fvqa} or proposed various fusion algorithms for multimodal information~\cite{marino2021krisp}. ConceptBERT~\cite{garderes2020conceptbert} constrains the multimodal information with question embedding and fuses embeddings of each modality for prediction. MAVEx~\cite{mavex} aims to discern the corresponding knowledge source for each candidate answer to reduce noise. KRISP~\cite{marino2021krisp} captures both implicit information in both questions, images and knowledge graphs.\\
\textbf{KVQA with LLMs.} Recently, large language models (LLMs) have surprised the community with their superior understanding of texts. PICa~\cite{pica} first leverages GPT3~\cite{gpt3} as an implicit knowledge source for reasoning by prompting the image captions and in-context examples. Another pipeline of studies employs LLMs to generate candidates or supporting evidence for particular captions, e.g., KAT~\cite{gui2022kat} and REVIVE~\cite{lin2022revive}. While they do not fully leverage the multiple sources of knowledge, 
we break the limitation of complex reasoning by developing a tailored multimodal fusion algorithm that balances intra- and inter-modal learning.

\section{Conclusions}
We present \texttt{MAIL}, a modality-aware integration with large language models for knowledge-based visual question answering. We formally define a novel multimodal learning paradigm for comprehensive cross-modal reasoning among multiple knowledge sources. The knowledge from LLMs is effectively leveraged via a carefully designed coupled graph construction, i.e., scene graph and concept graph. Then we integrate various multimodal information with a tailored pseudo-siamese graph medium fusion. It effectively enhances a tight inter-modal interaction and maximally preserves insightful intra-modal processing. \texttt{MAIL} achieves superiority on two benchmark datasets while possessing 24$\times$ less computational resources and 2$\sim 4\times$ faster inferential time than the existing state-of-the-art baselines.

\bibliography{MAIL}

\newpage
\newpage
\appendix
\section{Appendix}
\label{sec:appendix}
\subsection{Prompt Templates for Coupled Graph Construction}

\textbf{Prompt for Scene Graph Construction} \\
\textit{`Describe the image with as many details as possible. Generally, identify the objects and their spatial relations with each other. Specifically, include the visual outlook of different objects, e.g., color, style as well as the appearance for human beings.'} \\
\textbf{Prompt for Concept Graph Construction}\\
\textit{`Given the image caption, based on your comprehensive understanding, construct a high-quality scene graph with as many meaningful details of the mentioned entities as possible in the form of a triple (head entity, relation, tail entity). $\backslash n$ Strictly use the twelve predefined relations from: $\mathcal{R}$, e.g., \textit{(woman, in\_front\_of, car)}, \textit{(car, has\_color, blue)}, only return the triples with no other content. $\backslash n$ Caption: $\mathcal{D}$ $\backslash n$ Mentioned Entities: $\mathcal{M}$.'}

\subsection{Detailed Statistics of the Scene Graphs}
We showcase the beautiful distribution of the pre-defined condensed relations in the constructed scene graphs for OK-VQA and FVQA in Table~\ref{relations_ok}.

\subsection{Experiments}\label{basel}

{\bf Implementation Details}
We generate dense image captions with MiniGPT-4 (7B)~\cite{zhu2023minigpt}, and adopt ConceptNet~\cite{speer2018conceptnet} for external knowledge, one of the largest real-world commonsense KGs. We apply MiniGPT-4 with one Tesla V100. The entire processing of OK-VQA and the corresponding Microsoft COCO images~\cite{lin2014microsoft} including image-to-text and data cleaning takes about 4 rounds. We adopt $\ell=3$ and $\lambda=1e-3$ after hyperparameter tuning. The generated caption is stored for further multimodal learning. Our codes and processed graphs will be open-sourced and publicly available. 

For the results of baseline LLMs, since they could occasionally refuse to answer with responses like either `As a language model, I am not capable of understanding images' or `Sorry, there is no related information in the provided caption.', we report the average accuracy over 2 rounds.
\renewcommand{\arraystretch}{1.2}
\begin{table}[t!] \footnotesize
    
    \vspace{-3mm}
    \centering
    \setlength{\tabcolsep}{1mm}
    \begin{tabular}{c|c|c|c|c|c}
    \hline\hline
    \multirow{3}{*}{\textbf{Categories}} & \multirow{3}{*}{\textbf{Relation}}  & \multicolumn{2}{c}{\textbf{OK-VQA}} & \multicolumn{2}{c}{\textbf{FVQA}}\\
    \cmidrule{3-6}
    & & \textbf{Tain} & \textbf{Test}  & \textbf{Tain} & \textbf{Test}   
    \cr \cmidrule{1-6}
    \multirow{7}{*}{\makecell[c]{\textbf{Spatial}\\ \textbf{Features}}} & at\_location &10,562 &10,118 & 3,466 & 3,107\\
    &next\_to &3,948 & 3,772 & 2,533 & 2,289\\
    &in\_front\_of &2,239 & 2,244 & 759 & 687\\
    &surrounded\_by &2,004 & 2,026 & 699 & 549\\
    &covered\_by &180 & 191 & 9 & 7\\
    &includes &12,402 & 12,390 & 1,811 & 1,630\\
    &holds &3,344 & 3,090 & 965 & 794
    \cr \cmidrule{1-6}
    \multirow{5}{*}{\makecell[c]{\textbf{Object}\\ \textbf{Features}}} & has\_property &16,685 &17,032 & 1,301 & 1,297\\
    &has\_color &9,191 & 8,836 & 3,653 & 3,258\\
    &made\_of &3,388 & 3,310 & 978 & 948\\
    &wears &5,172 & 5,049 & 1,504 & 1,449\\
    &intends\_to &1,599 & 1,655 & 9 & 8
    \cr \cmidrule{1-6}\hline
    \end{tabular}    
    \caption{The overall statistics of the pre-defined condense relations for OK-VQA and FVQA datasets. They depict the spatial features and object features in images.}
    \label{relations_ok}
    \vspace{-4mm}
\end{table}
{\bf Baselines}
Specifically, for traditional end-to-end baselines, we pick the representative state-of-the-art methods, i.e., a direct answering based on questions only (Q Only)~\cite{marino2019ok},BAN~\cite{kim2018bilinear}, MUTAN~\cite{ben2017mutan}, ConceptBERT~\cite{garderes2020conceptbert}, KRISP~\cite{marino2021krisp}, MAVEx~\cite{mavex}, VLC-BERT~\cite{ravi2023vlc}, HCNMN~\cite{zhang2023toward} and MCAN~\cite{MCAN}. Moreover, as BAN and MUTAN merely learn the uni-modal visual features, they are augmented with ArticleNet (AN)~\cite{marino2019ok} that is trained to retrieve knowledge from Wikipedia for corresponding question-image pair to facilitate the reasoning with external knowledge, denoted as `BAN + AN' and `MUTAN + AN'~\cite{marino2019ok}. 

While for LLM-enhanced baselines, we adopt PICa~\cite{pica}, KAT~\cite{gui2022kat}, and REVIVE~\cite{lin2022revive}.

\subsection{Generalization on FVQA Dataset}\label{FVQA}
To further demonstrate the generalization ability of our proposed \texttt{MAIL}, we compare it with the widely adopted baselines on the first KVQA dataset \textbf{FVQA}, i.e., XNM~\cite{XNM}, KI-Net~\cite{KI-Net}, UnifER~\cite{unifer}, MCAN~\cite{MCAN} and HCNMN~\cite{zhang2023toward}. For external knowledge, KI-Net uses ConceptNet and Wordnet; UnifER uses VisualBert, LXMERT and ViLT; HCNMN uses WordNet, WikiText, ConceptNet and Visual Genome.

\end{document}